\documentclass{article}
\usepackage{ijcai18}

\usepackage{times}
\usepackage{xcolor}
\usepackage{soul}
\usepackage[utf8]{inputenc}
\usepackage[small]{caption}

\usepackage{latexsym}
\usepackage{amsmath}
\usepackage{amsfonts}
\usepackage{multirow}
\usepackage{makecell}
\usepackage{graphicx}
\usepackage{bm}

\title{Chinese Poetry Generation with a Working Memory Model}

\author{
Xiaoyuan Yi$^1$, 
Maosong Sun$^1$\thanks{Corresponding author: M. Sun (sms@mail.tsinghua.edu.cn)}, 
Ruoyu Li$^2$, 
Zonghan Yang$^1$
\\ 
$^1$ State Key Lab on Intelligent Technology and Systems, \\
Beijing National Research Center for Information Science and Technology, \\  Department of Computer Science and Technology, Tsinghua University, Beijing, China \\
$^2$ 6ESTATES PTE LTD, Singapore\\
}
\begin{document}

\maketitle

\begin{abstract}
As an exquisite and concise literary form, poetry is a gem of human culture. Automatic poetry generation is an essential step towards computer creativity. In recent years, several neural models have been designed for this task. However, among lines of a whole poem, the coherence in meaning and topics still remains a big challenge. In this paper, inspired by the theoretical concept in cognitive psychology, we propose a novel Working Memory model for poetry generation. Different from previous methods, our model explicitly maintains topics and informative limited history in a neural memory. During the generation process, our model reads the most relevant parts from memory slots to generate the current line. After each line is generated, it writes the most salient parts of the previous line into memory slots. By dynamic manipulation of the memory, our model keeps a coherent information flow and learns to express each topic flexibly and naturally. We experiment on three different genres of Chinese poetry: quatrain, iambic and chinoiserie lyric. Both automatic and human evaluation results show that our model outperforms current state-of-the-art methods.
\end{abstract}
\section{Introduction}
Poetry is a literary form with concise language, exquisite expression and rich content, as well as some structural and phonological requirements. During the thousands of years of human history, poetry is always fascinating and popular, influencing the development of different countries, nationalities and cultures. 
\begin{figure}
\centering
\includegraphics[scale=0.30]{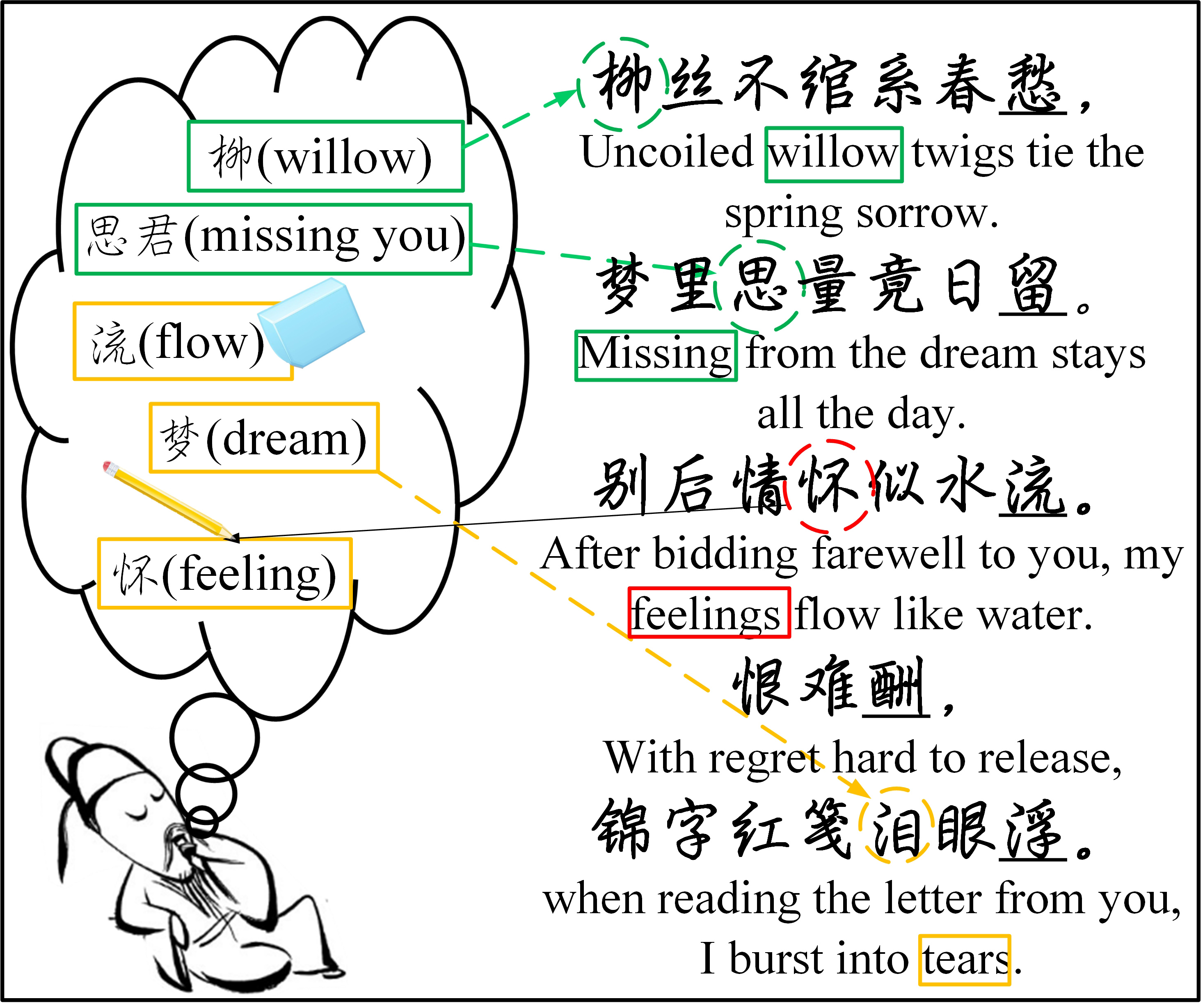}
\caption{An iambic generated by our model with the tune \emph{Remember the Prince}, taking \emph{liu} (willow) and \emph{si jun} (missing you) as input topic words. Rhyming characters are underlined. The left part is an artistic illustration of our model, where solid and dotted arrows represent memory writing and reading respectively.}
\label{fig1}
\end{figure}

In Chinese, there are different genres of poetry. In this work, we mainly focus on three of them: quatrain (\emph{Jueju}), iambic (\emph{Ci}) and chinoiserie lyric. Both for quatrain and iambic, there are various tunes (sub-genres) and each tune defines the length of each line, the tone of each character and the number of lines (for iambic). With more than eight hundred tunes, iambic is a quite complex genre (as shown in Figure \ref{fig1}). By contrast, chinoiserie lyric is relatively free except for the requirement on rhyme, which gets popular in recent twenty years, driven by some famous singers \cite{Fung:07}.

We concentrate on automatic poetry generation. Besides the requirements on form, to create a high-quality poem, how to achieve better coherence is a key problem across different genres. Generally, two factors must be taken into account. For one thing, the topic needs to be expressed in a poem flexibly. For multiple topics, natural transition among different topics can improve coherence. For another, lines in a poem should be coherent in meaning, theme and artistic conception.

Recently, several neural models have been designed for different aspects of this task, such as poetry style transfer \cite{Zhang:17} and rhythmic constraints \cite{Marjan:16}. Nevertheless, this fundamental problem, coherence, hasn't been handled well, which is a major reason for the gap between computer-generated poems and the human-authored ones. The key point lies in that when generating a poem line, existing models assume user inputs (topics) and the history (preceding generated lines in the poem) can be packed into a \emph{single} small vector \cite{Yan:16}, or assume the model is able to focus on the most important part of an \emph{unlimited} history \cite{Wang:16b}, which are implausible and against a human writing manner.

To tackle this problem, we refer to the concept in cognitive psychology, where the working memory is a system with a \emph{limited} capacity that is responsible for holding information available for reasoning, decision-making and behaviour \cite{Shah:99}. Previous work has demonstrated the importance of working memory in writing \cite{McCutchen:00}. From the perspective of psycholinguistics, coherence is achieved if the reader can connect the incoming sentence to the content in working memory and to the major messages and points of the text \cite{Sanders:01}.

Inspired by this, we propose a novel Working Memory model\footnote{In fact, part of our model can be also considered as a kind of Neural Turing Machine \cite{Graves:14}. We take the perspective of working memory here to emphasize the influence of this structure on human writing.} for poetry generation. Rather than merges user topic words as one vector as previous work \cite{Yan:16}, our model maintains them in the memory explicitly and independently, which play the role of `major messages'. When generating each line, our model learns to read most relevant information (topics or history) from the memory to guide current generation, according to what has been generated and which topics have been expressed so far. For each generated line, our model selects the most salient parts, which are informative for succeeding generation, and writes them into the memory. Instead of full history, our model keeps informative partial history in \emph{multiple} but \emph{limited} memory slots. This dynamical reading-and-writing way endows the model with the ability to focus on relevant information and to ignore distractions during the generation process, and therefore improves coherence to a significant extent. Besides, we design a special genre embedding to control the tonal category of each character and the length of each line, which makes our model structure-free and able to generate various genres of poetry.

In summary, the contributions of this paper are as follows:
\begin{itemize}
\item To the best of our knowledge, for poetry generation, we first propose to exploit history with a dynamically reading-and-writing memory. 
\item We utilize a special genre embedding to flexibly control the structural and phonological patterns, which enables our model to generate various genres of poetry.
\item On quatrain, iambic and chinoiserie lyric, our model outperforms several strong baselines and achieves new state-of-the-art performance.
\end{itemize}
\section{Related Work}
As a long-standing concern of AI, the research on automatic poetry generation can be traced back decades. The first step of this area is based on rules and templates \cite{Gervas:01}. Since the 1990s, statistical machine learning methods are adopted to generate poetry, such as genetic algorithms \cite{Manurung:03} and statistical machine translation (SMT) approach \cite{He:12}.

Stepping into the era of neural networks, different models have been proposed to generate poetry and shown great advantages. In general, previous neural models fall under three methodologies in terms of how the history (preceding generated lines in the poem) is exploited.

The first methodology is to pack all history into a \emph{single} history vector. Zhang and Lapata first \shortcite{Zhang:14} propose to generate Chinese quatrains with Recurrent Neural Network (RNN). Each generated line is vectorized by a Convolutional Sentence Model and then packed into the history vector. To enhance coherence, their model needs to be interpolated with two extra SMT features, as the authors state. Yan \shortcite{Yan:16} generates Chinese quatrains using two RNNs. The last hidden state in the first RNN is used as the line vector, which is packed into a history vector by the second RNN. In his model, the poem generated in one pass will be refined for several times with an iterative polishing schema to improve quality.

The second one is to concatenate full history as a long sequence, which is exploited by a sequence-to-sequence model with attention mechanism \cite{Bahdanau:15}. \cite{Wang:16a} proposes a two-stage Chinese quatrains generation method which plans sub-keywords of the poem in advance by a language model, then generates each line with the aid of the planned sub-keyword. However, such planning of keywords takes a risk of losing flexibility in topic expression.

The last one is to take the whole poem as a long sequence and generate it word by word, where history propagates implicitly along the RNN. This methodology is used to generate both English poetry \cite{Hopkins:17,Ghazvininejad:17} and Chinese poetry \cite{Zhang:17,Wang:16b}

These neural network-based approaches are promising, but there is still a lot of room for improvement. A single vector doesn't have enough capacity to maintain the full history. Moreover, informative words and noises (e.g., stop words) are mixed, which hinders the exploitation of history. When the input or the output sequence is too long, the performance of sequence-to-sequence model will still degrade, even with an attention mechanism, which has been observed in related tasks, e.g., Neural Machine Translation \cite{Shen:16}. Consequently, we propose our Working Memory model with \emph{multiple} but \emph{limited} memory slots.

Memory Network (MN) \cite{Weston:15} and Neural Turing Machine (NTM) have shown great power in some tasks, e.g., Question Answering (QA). The most relevant work to our model is \cite{Zhang:17}, which saves hundreds of human-authored poems in a \emph{static} external memory to improve the innovation of generated quatrains and achieve style transfer. In fact, these MN and NTM models just learn to write external texts (poems or articles) into memory. By contrast, our model writes the generated history and hence adopts a dynamic utilization of memory, which is closer to a human manner as discussed in Section 1.
\section{Model Description}
\subsection{Overview}
\begin{figure}
\centering
\includegraphics[scale=0.24]{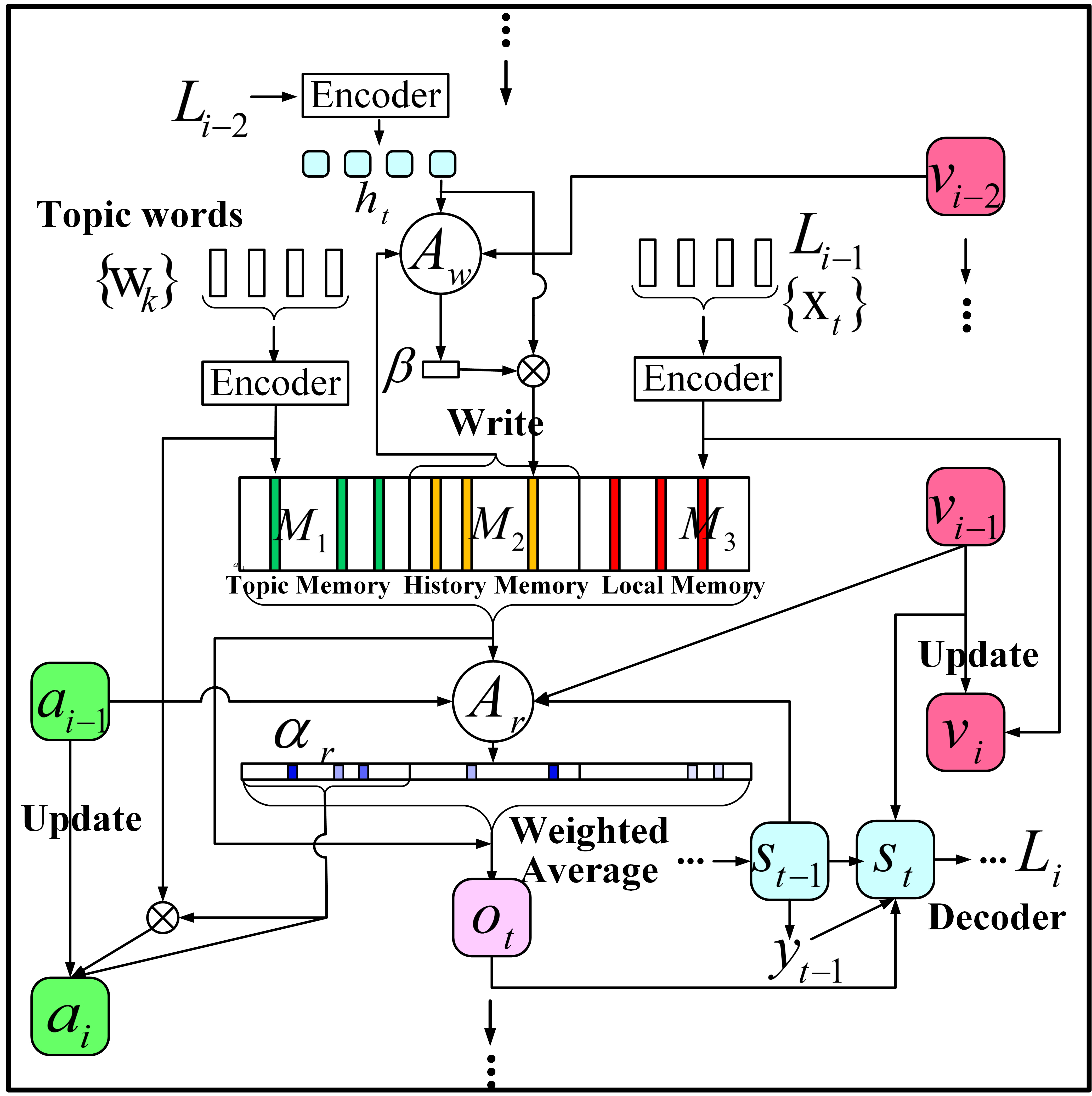}
\caption{A graphical illustration of the Working Memory model, which consists of an encoder, a decoder and the working memory. The top half of this figure shows memory writing before generating $L_i$, and the bottom half shows the generation of $L_i$.}
\label{fig2}
\end{figure}
Before presenting the proposed model, we first formalize our task. The inputs are user topics specified by $K_1$ keywords, $\{w_k\}_{k=1}^{K_1}$. The output is a poem consisting of n lines, $\{L_i\}_{i=1}^{n}$. Since we take the sequence-to-sequence framework and generate a poem line by line, the task can be converted to the generation of an i-th line which is coherent in meaning and related to the topics, given previous i-1 lines $L_{1:i-1}$ and the topic words $w_{1:K_1}$.

As illustrated in Figure \ref{fig2}, the working memory is comprised of three modules: topic memory $M_1 \in \mathbb{R}^{K_1*d_h}$, history memory $M_2 \in \mathbb{R}^{K_2*d_h}$ and local memory $M_3 \in \mathbb{R}^{K_3*d_h}$, where each row of the matrices is a memory slot and $d_h$ is slot size. $K_2$ and $K_3$ are the numbers of slots. Therefore the whole working memory $M$ = $[M_1;M_2;M_3]$, $M \in \mathbb{R}^{K*d_h}$ where $[;]$ means concatenation and $K$=$K_1$+$K_2$+$K_3$.

Each topic word $w_k$ is written into the topic memory in advance, which acts as the `major message' and remains unchanged during the generation process of a poem. Before generating the i-th line $L_i$, each character of $L_{i-1}$ is written into the local memory. There are often strong semantic associations between two adjacent lines in Chinese poetry, therefore we feed $L_{i-1}$ into this local memory to provide full short-distance history. Different from other two modules, the model selects some salient characters of $L_{i-2}$ to write into the history memory. In this way, the history memory maintains informative partial long-distance history. These three modules are read jointly. 

Following this procedure, we detail our model.
\subsection{Working Memory Model}
Based on the sequence-to-sequence framework, we use GRU \cite{Cho:14} for decoder and bidirectional encoder. Denote X a line in encoder ($L_{i-1}$), $X=(x_1x_2 \ldots x_{T_{enc}})$, and Y a generated line in decoder ($L_i$), $Y=(y_1y_2 \ldots y_{T_{dec}})$. $h_t$ and ${s_t}$ represent the encoder and decoder hidden states respectively. $e(y_t)$ is the word embedding of $y_t$. The probability distribution of each character to be generated in $L_i$ is calculated by\footnote{For brevity, we omit biases and use $h_t$ to represent the combined state of bidirectional encoder.}:
\begin{align}
& s_t = GRU(s_{t-1}, [e(y_{t-1}); o_t; g_t; v_{i-1}]), \\
& p(y_t|y_{1:t-1},L_{1:i-1},w_{1:K_1}) = softmax(Ws_t),
\end{align}

where $o_t$ is the memory output and W is the projection parameter. $v_{i-1}$ is a global trace vector, which records what has been generated so far and provides implicit global information for the model. Once $L_i$ is generated, it is updated by a simple vanilla RNN:
\begin{align}
& v_i = \sigma(v_{i-1}, \frac{1}{T_{enc}}\sum_{t=1}^{T_{enc}} h_t), v_0=\bm{0}.
\end{align}

$\sigma$ defines a non-linear layer and $\bm{0}$ is a vector with all 0-s.

\textbf{Genre embedding}. $g_t$ in Eq. (1) is a special genre embedding. Since poetry must obey structural and phonological rules, we use this genre embedding to control the genre of a generated poem over each character. $g_t$ is the concatenation of a phonology embedding and a length embedding, which are learned during training. We define 36 phonology categories in terms of \cite{Ge:09}. The phonology embedding indicates the required category of $y_t$. The length embedding indicates the number of characters to be generated after $y_t$ in $L_i$ and hence controls the length of $L_i$. 

\textbf{Memory Reading}. We begin by defining an \emph{Addressing Function}, $\alpha = A(\tilde{M}, q)$, which calculates the probabilities that each slot of the memory is to be selected and operated. Concretely, we have:
\begin{align}
& z_{k} = b^T \sigma(\tilde{M}[k], q), \\
& \alpha[k] = softmax(z_{k}),
\end{align}

where $q$ is the query vector, $b$ is the parameter, $\tilde{M}$ is the memory to be addressed, $\tilde{M}[k]$ is the k-th slot (row) of $\tilde{M}$ and $\alpha[k]$ is the k-th element in vector $\alpha$.  

Then, the working memory is read as:
\begin{align}
& \alpha_r = A_r(M, [s_{t-1}; v_{i-1}]),\\
& o_t = \sum_{k=1}^K \alpha_r[k]*M[k],
\end{align}

where $\alpha_r$ is the reading probability vector and the trace vector $v_{i-1}$ is used to help the Addressing Function avoid reading redundant content. Joint reading from the three memory modules enables the model to flexibly decide to express a topic or to continue the history content.

\textbf{Memory Writing}. Here we use hidden states as vector representations of characters. For topic memory, we feed characters of each topic word $w_k$ into the encoder, then get a topic vector by a non-linear transformation of the corresponding hidden states. Then each topic vector is directly filled into a slot. Before generating $L_i$, the encoder hidden states of characters in $L_{i-1}$ are filled into local memory slots.

After $L_i$ is generated and before the generation of $L_{i+1}$, for each encoder state $h_t$ of $L_{i-1}$, the model select a history memory slot by writing addressing function and fill $h_t$ into it. Formally, we have:
\begin{align}
& \alpha_w = A_w(\tilde{M_2}, [h_t;v_{i-1}]),  \\
& \beta[k] = I(k=\mathop{\arg\max}_j \alpha_w[j]), \\
& \tilde{M}_2[k] \leftarrow (1-\beta[k]) * \tilde{M}_2[k] + \beta[k] * h_t,
\end{align}

where $I$ is an indicator function and $\alpha_w$ is the writing probabilities vector. $\tilde{M}_2$ is the concatenation of history memory $M_2$ and a null slot. If there is no need to write $h_t$ into history memory, model learns to write it into the null slot, which is ignored when reading memory by Eq. (6).

Since Eq. (9) is non-differentiable, it is only used for testing. For training, we simply approximate $\beta$ as:
\begin{align}
& \beta = tanh(\gamma * (\alpha_w - \bm{1} * max (\alpha_w) )) + \bm{1},
\end{align}

where $\bm{1}$ is a vector with all 1-s and $\gamma$ is a large positive number. Eq. (11) is a rough approximation but it's differentiable, by which the model learns to focus on one slot with a higher writing probability. We expect $h_t$ to be written into only one slot, because we want to keep the representations of salient characters independent as discussed in Section 2.

Before the generation, all memory slots are initialized with $\bm{0}$. For empty slots, a random bias is added to $z_k$ in Eq. (5) to prevent multiple slots getting the same probability.
\subsection{Topic Trace Mechanism}
Though we use a global trace vector $v_i$ to save all generated content, it seems not enough to help the model remember whether each topic has been used or not. Therefore we design a \textbf{Topic Trace (TT)} mechanism, which is a modified coverage model \cite{Tu:16}, to record the usage of topics in a more explicit way:
\begin{align}
& c_i = \sigma(c_{i-1}, \frac{1}{K_1} \sum_{k=1}^{K_1} M[k] * \alpha_r[k]),c_0=\bm{0},  \\
& u_i = u_{i-1} + \alpha_{r}[1:K_1], u_i \in \mathbb{R}^{K_1*1}, u_0=\bm{0}, \\
& a_i = [c_i;u_i].
\end{align}

$c_i$ maintains the content of used topics and $u_i$ explicitly records the times of reading each topic. $a_i$ is the topic trace vector. Then we rewrite Eq. (6) as:
\begin{align}
\alpha_r = A_r(M, [s_{t-1}; v_{i-1}; a_{i-1}]).
\end{align}

We will show that this Topic Trace mechanism can further improve the performance of our model in Section 4.
\begin{table}
\centering
\begin{tabular}{|l|l|l|l|}
\Xhline{1.2pt}
 & \# of Poems & \# of Lines & \# of Characters  \\
\Xhline{1.2pt}
Quatrains & 72,000 & 288,000 & 1,728,000 \\
\hline
Iambics & 33,499 & 418,896 & 2,099,732\\
\hline
Lyrics & 1,079 &  37,237 & 263,022 \\
\Xhline{1.2pt}
\end{tabular}
\caption{Details of our corpus.}
\label{tab1}
\end{table}
\section{Experiments}
\subsection{Data and Setups}
Table \ref{tab1} shows details of our corpus. We use 1,000 quatrains, 843 iambics and 100 lyrics for validation; 1,000 quatrains, 900 iambics and 100 lyrics for testing. The rest are used for training. 

Since our model and most previous models take topic words as input, we run TextRank \cite{Mihalcea:2004} on the whole corpus and extract four words from each poem. In training, we build four $<$keyword(s), poem$>$ pairs for each poem using 1 to 4 keywords respectively, so as to improve the model's ability to cope with different numbers of keywords. In testing, we randomly select one pair from the four and use the phonological and structural pattern of ground truth.

We set $K_1$ = 4 and $K_2$ = 4. The sizes of word embedding, phonology embedding, length embedding, hidden state, global trace vector, topic trace vector are set to 256, 64, 32, 512, 512, 24 (20+4) respectively. Since we directly feed hidden states of bidirectional encoder into memory, the slot size $d_h$ is 1024. The word embedding is initialized with word2vec vectors pre-trained on the whole corpus. Different memory modules share the same encoder. We use two different addressing functions for reading and writing respectively. For all non-linear layers, tanh is used as the activation function.

Adam with shuffled mini-batches (batch size 64) is used for optimization. To avoid overfitting, 25\% dropout and $\ell_2$ regularization are used. Optimization objective is standard cross entropy errors of the predicted character distribution and the actual one. Given several topic words as input, all models generate each poem with beam search (beam size 20). For fairness, all baselines share the same configuration.
\subsection{Models for Comparisons}
\begin{table}
\centering
\begin{tabular}{|c||c|c|c|}
\Xhline{1.2pt}
& \bf Models & \bf BLEU & \bf PP \\
\Xhline{1.2pt}

\multirow{2}{*}{\bf Quatrains} & iPoet & 0.425 & 138\\ 
& WM & \bf 1.315 & \bf 86 \\
\hline

\multirow{2}{*}{\bf Iambics} & iambicGen & 0.320 & 262 \\
& WM & \bf 0.699 &  \bf 72  \\
\hline

\multirow{2}{*}{\bf Lyrics} & lyricGen & 0.312 & 302 \\
& WM & \bf 0.568 & \bf 138 \\
\Xhline{1.2pt}
\end{tabular}
\caption{Automatic evaluation results. BLEU scores are calculated by the multi-bleu.perl script. PP means perplexity.}
\label{tab2}
\end{table}
\begin{table}
\small
\centering
\begin{tabular}{|c|c|c|c|c|c|c|}
\Xhline{1.2pt}
\multirow{2}{*}{\bf Strategies} & \multicolumn{2}{c|}{ \bf Quatrains} & \multicolumn{2}{c|}{\bf Iambics} & \multicolumn{2}{c|}{\bf Lyrics} \\
\cline{2-7}
& BLEU & PP & BLE & PP & BLEU & PP \\
\Xhline{1.2pt}

WM$_0$ & 1.019 & 127 & 0.561 & 152 & 0.530 & 249\\ 
WM$_0$+GE & 1.267 & 87 & 0.672 & 74 & 0.542 & 144\\ 
WM$_0$+GE+TT & \bf 1.315 & \bf 86 & \bf 0.699 & \bf 72 & \bf 0.568 & \bf 138\\ 
\Xhline{1.2pt}
\end{tabular}
\caption{Comparison of different strategies. GE: genre embedding. TT: Topic Trace mechanism. WM$_0$ is the model without GE or TT.}
\label{tab3}
\end{table}
\begin{table*}[htp]
\centering
\begin{tabular}{|c|c|l|l|l|l|l|}

\Xhline{1.2pt}
 & \bf Models & \bf Fluency & \bf Meaning & \bf Coherence & \bf Relevance & \bf Aesthetics \\
\Xhline{1.2pt}

\multirow{5}{*}{\bf Quatrains} & Planning & 2.28 & 2.13 & 2.18 & 2.50 & 2.31 \\ 
& iPoet & 2.54 & 2.28 & 2.27 & 2.13 & 2.45 \\
& FCPG & 2.36 & 2.15 & 2.15 & 2.65 & 2.28 \\
& WM & \bf 3.57$^{**}$ & \bf 3.45$^{**}$ & \bf 3.55$^{**}$ & \bf 3.77$^{**}$ & \bf 3.47$^{**}$ \\
& Human & 3.62 & 3.52 & 3.59 & 3.78 & 3.58 \\

\Xhline{1.2pt}

\multirow{3}{*}{\bf Iambics} & iambicGen & 2.48 & 2.73 & 2.78 & 2.36 & 3.08 \\
& WM & \bf 3.39$^{**}$ & \bf 3.69$^{**}$ & \bf 3.77$^{**}$ & \bf 3.87$^{**}$ & \bf 3.87$^{**}$ \\
& Human & 4.04 & 4.10$^{++}$ & 4.13$^{++}$ & 4.03 & 4.09 \\
\Xhline{1.2pt}

\multirow{3}{*}{\bf Lyrics} & lyricGen & 1.70 & 1.65 & 1.81 & 2.24 & 1.99 \\
& WM & \bf 2.63$^{**}$ & \bf 2.49$^{**}$ & \bf 2.46$^{**}$ & \bf 2.53 & \bf 2.66$^{**}$ \\
& Human & 3.43$^{++}$ & 3.20$^{++}$ & 3.41$^{++}$ & 3.34$^{++}$ & 3.26$^{++}$ \\
\Xhline{1.2pt}
\end{tabular}
\caption{Human evaluation results. Diacritic ** ($p < 0.01$) indicates WM significantly outperforms baselines; ++ ($p < 0.01$) indicates Human is significantly better than all models. The Intraclass Correlation Coefficient of the four groups of scores is 0.5, which indicates an acceptable inter-annotator agreement.}
\label{tab4}
\end{table*}
Besides \textbf{WM}\footnote{https://github.com/xiaoyuanYi/WMPoetry.} (our Working Memory model) and \textbf{Human} (human-authored poems), on quatrains we compare \textbf{iPoet} \cite{Yan:16}, \textbf{Planning} \cite{Wang:16a} and \textbf{FCPG} \cite{Zhang:17}. We choose these previous models as baselines, because they all achieve satisfactory performance and the authors have done thorough comparisons with other models, such as RNNPG \cite{Zhang:14} and SMT \cite{He:12}. Moreover, the three models just belong to the three methodologies mentioned in Section 2 respectively.

On iambics we compare \textbf{iambicGen} \cite{Wang:16b}. To the best of our knowledge, this is the only one neural model designed for Chinese iambic generation.

On chinoiserie lyrics, since there is no specially designed model in the literature, we implement a standard sequence-to-sequence model as the baseline, called \textbf{lyricGen}.
\subsection{Evaluation Design}
\textbf{Automatic Evaluation}. Referring to \cite{Zhang:14,Yan:16}, we use BLEU and perplexity to evaluate our model. BLEU and Perplexity are not perfect metrics for generated poems, but they can still provide an aspect for evaluation and make sense to some extent in the context of pursuing better coherence. Furthermore, automatic evaluation can save much labour and help us determine the best configure.

\textbf{Human Evaluation}. We design five criteria: \textbf{Fluency} (does the poem obey the grammatical, structural and phonological rules?), \textbf{Meaning} (does the poem convey some certain messages?), \textbf{Coherence} (is the poem as a whole coherent in meaning and theme?), \textbf{Relevance} (does the poem express user topics well?), \textbf{Aesthetics} (does the poem have some poetic and artistic beauties?). Each criterion needs to be scored in a 5-point scale ranging from 1 to 5.

From the testing set, for quatrains, iambics and lyrics we randomly select 30, 30 and 20 sets of topic words respectively to generate poems with these models. For Human, we select poems containing the given words. Therefore, we obtain 150 quatrains (30*5), 90 iambics (30*3) and 60 lyrics (20*3). We invite 16 experts\footnote{The experts are Chinese literature students or members of a poetry association. They are required to focus on the quality as objectively as possible, even if they recognize the human-authored ones.} on Chinese poetry to evaluate these poems, who are divided into four groups. Each group completes the evaluation of all poems and we use the average scores.

Planning and FCPG are not suitable for automatic evaluation, because FCPG is designed for innovation and Planning will plan the sub-topics by itself, which increase the perplexity. Thus we leave them for human evaluation.
\subsection{Evaluation Results}
As shown in Table \ref{tab2}, WM outperforms other models under BLEU and perplexity evaluations. On quatrains, WM gets almost three times higher BLEU score than iPoet does. This significant improvement partially lies in that more than 70\% of the input topics are expressed\footnote{If a topic word or at least one of its relevant words is generated, we say this topic is expressed.} in poems generated by WM, benefiting from the topic memory. By contrast, this expression ratio is only 28\% for iPoet, since iPoet merges words and history into two single vectors respectively, resulting in implicit and indiscriminate exploitation of topics and history. On iambics, WM also achieves notable performance. Because iambicGen generates the whole iambic as a long sequence by the decoder, it handles short iambics well but fails to generate high-quality longer ones. For iambics with less than 70 characters, perplexity of iambicGen is 235. For those with more characters, perplexity of iambicGen increases to 290. On chinoiserie lyrics, WM also gets better results, though the performance is not so satisfactory (both for WM and lyricGen), due to the small training set.

It is worth mentioning that the improvement partially results from the genre embedding. By incorporating structural and phonological control into the model, WM greatly reduces the uncertainty of generation. To demonstrate the effectiveness of the working memory itself, we show the performance of different strategies of WM in Table \ref{tab3}. As we can see, even without genre embedding, our model still outperforms baselines prominently. Besides, Topic Trace mechanism further improves performance.
\begin{figure*}
\centering
\includegraphics[scale=0.33]{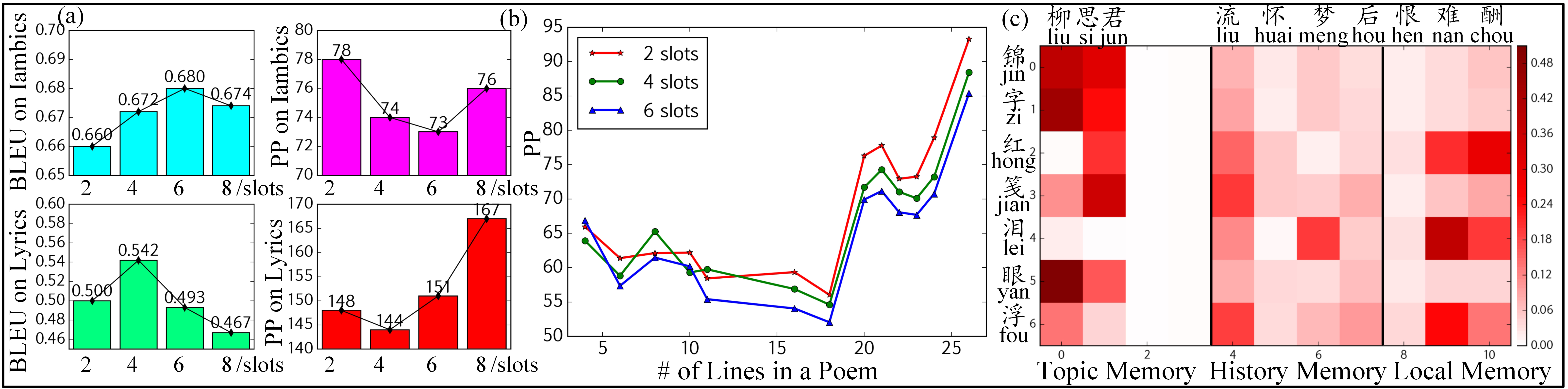}
\caption{(a) Over different numbers of history memory slots, BLEU and perplexity on iambics and lyrics. (b) On iambics, perplexity over different numbers of lines in a poem. (c) The visualization of memory (in the x-axis) reading probabilities, $\alpha_r$, when generating the last line (in the y-axis) of the iambic shown in Figure \ref{fig1}.}
\label{fig3}
\end{figure*}

Table \ref{tab4} gives human evaluation results. WM achieves better results than other models. On quatrains, WM gets close to Human on Coherence and Relevance. Planning gets the worst results on Fluency and Meaning. This is mainly because planning mechanism can't guarantee the quality of planned sub-keywords and the fixed keywords order loses some freedom of topic expression, hurting fluency and meaning. iPoet gets the lowest score on Relevance, since it packs all topic words into one vector, resulting in a low topic expression ratio. By contrast, WM maintains keywords in the topic memory independently and the expression order is flexibly decided by the model in terms of the history. Benefiting from TT, an unexpressed word still has the chance to be generated in later lines. Thus WM gets a comparable score with Human on Relevance. FCPG performs worst on Coherence. As discussed in Section 2, FCPG generates the whole poem as a long sequence and the history is saved in RNN state implicitly, which therefore can't be utilized effectively. On iambics and lyrics, WM gets better results, but there is still a distinct gap with Human. Iambic is a quite complex form and the longest iambic in our testing set consists of more than 150 characters (25 lines). It's much harder for the model to generate a high-quality iambic. For lyrics, due to the limited small data, the results are not as good as we expected. We put the requirements of structure and phonology into Fluency criterion. As a result, WM gets a much higher Fluency score than baselines, benefiting from the genre embedding.
\subsection{Analyses and Discussions}
We test the performance of WM\footnote{We removed Topic Trace here to observe the influence of the number of slots itself.} on different numbers of slots. As shown in Figure \ref{fig3} (a), both on iambics and lyrics, as the number of slots increases, BLEU gets better first and then deteriorates and so does perplexity. Some lyrics consist of more than 100 lines. More slots should have led to better results on lyrics. However, with the small lyrics corpus, the model can't be trained adequately to operate many slots. Figure \ref{fig3} (b) gives perplexity over different numbers of lines on iambics. There is little difference for iambics with less than 10 lines. For longer iambics, the model with 6 slots gets better results, though perplexity still increases with more lines.

With too many slots (e.g., infinite slots), our history memory falls back to the second methodology discussed in Section 2. Without any slot, it falls back to the first methodology. The number of memory slots is an important parameter and should be balanced carefully in accordance with the conditions.

In Figure \ref{fig3} (c), we show an example of how our model focuses on different parts of the memory when generating a line. Our model ignores topic word \emph{liu} (willow) when generating character \emph{hong} (red), since the color of willow is green. The model focuses on topic word \emph{si jun} (missing you) when generation character \emph{jian} (letter), since in ancient China, people often sent their love and missing by letters. Besides, the model generates \emph{lei} (tears) with a strong association with \emph{meng} (dream) in history memory. The word `dream' is often a symbol to express the pain that a girl is separated from her lover and can only meet him in the dream.
\section{Conclusion and Future Work}
In this paper, we address the problem of pursuing better coherence in automatic poetry generation. To this end, a generated poem as a whole should be relevant to the topics, express these topics naturally and be coherent in meaning and theme. Inspired by the concept in cognitive psychology, we propose a Working Memory model, which maintains user topics and informative limited history in memory to guide the generation. By dynamical reading and writing during the generation process, our model keeps a coherent information flow and ignores distractions. The experiment results on three different genres of Chinese poetry demonstrate that our model effectively improves the quality and coherence of generated poems to a significant extent.

Besides, combined with a genre embedding, our model is able to generate various genres of poetry. The specially designed Topic Trace mechanism helps the model remember which topics have been used in a more explicit way, further improving the performance.

There still exists a gap between our model and human poets, which indicates that there are lots to do in the future. We plan to design more effective addressing functions and incorporate external knowledge to reinforce the memory.
\section*{Acknowledgments}
This work is supported by the National 973 Program (No.2014CB340501).

\clearpage

\bibliographystyle{named}
\bibliography{ijcai18}

\end{document}